\UseRawInputEncoding

\documentclass[letterpaper, 10 pt, conference]{ieeeconf}  

\IEEEoverridecommandlockouts                              

\usepackage{graphics} 
\usepackage{epsfig} 
\usepackage{mathptmx} 
\usepackage{times} 
\usepackage{amsmath} 
\usepackage{amssymb}  
\usepackage{graphicx}
\usepackage{stfloats}
\usepackage{diagbox}
\usepackage{url}
\usepackage{cite}
\usepackage{array}
\usepackage{graphbox}
\usepackage[caption=false]{subfig} 
\usepackage{booktabs}
\usepackage[colorlinks,linkcolor=black]{hyperref}

\begin{document}

\title{\LARGE \bf Decoupling Skill Learning from Robotic Control for Generalizable Object Manipulation}

\author{Kai Lu$^{1}$, Bo Yang$^{2*}$, Bing Wang$^{1}$, Andrew Markham$^{1}$
\thanks{$^{1}$: K. Lu, B. Wang, and A. Markham are with the Department of Computer Science, University of Oxford, Oxford, UK.
        {\tt\small \{kai.lu, bing.wang, andrew.markham\}@cs.ox.ac.uk}}%
\thanks{$^{2}$: B. Yang is with vLAR Group, Department of Computing, Hong Kong Polytechnic University, HKSAR.
        {\tt\small bo.yang@polyu.edu.hk}}%
\thanks{$^{*}$: Corresponding author.}
}

\maketitle
\thispagestyle{empty}
\pagestyle{empty}

\begin{abstract}
Recent works in robotic manipulation through reinforcement learning (RL) or imitation learning (IL) have shown potential for tackling a range of tasks e.g., opening a drawer or a cupboard. However, these techniques generalize poorly to unseen objects. We conjecture that this is due to the high-dimensional action space for joint control. In this paper, we take an alternative approach and separate the task of learning `what to do' from `how to do it' i.e., whole-body control. We pose the RL problem as one of determining the skill dynamics for a disembodied virtual manipulator interacting with articulated objects. The whole-body robotic kinematic control is optimized to execute the high-dimensional joint motion to reach the goals in the workspace. It does so by solving a quadratic programming (QP) model with robotic singularity and kinematic constraints. Our experiments on manipulating complex articulated objects show that the proposed approach is more generalizable to unseen objects with large intra-class variations, outperforming previous approaches. The evaluation results indicate that our approach generates more compliant robotic motion and outperforms the pure RL and IL baselines in task success rates. Additional information and videos are available at \href{https://kl-research.github.io/decoupskill}{\textcolor{blue}{https://kl-research.github.io/decoupskill}}.

\end{abstract}

\section{INTRODUCTION}

Robotic manipulation has a broad range of applications, such as industrial automation, healthcare, and domestic assistance. For repetitive tasks in controlled environments, e.g., automotive assembly, robotic manipulation has enjoyed many years of success. However, commonly used methods, such as model-predictive control and off-the-shelf planners, typically require accurate physical models of objects and the environment, which are largely unavailable in uncontrolled settings `in-the-wild'. 
Learning-based methods have recently been studied in household manipulation tasks \cite{mu_maniskill_2021, szot_habitat_2021, ehsani_manipulathor_2021, gan_threedworld_2021}, where a visual control policy is learned either from interactions via reinforcement learning (RL) or from demonstrations via imitation learning (IL). 
However, a common concern of RL is the unproductive exploration in the high-dimensional continuous action space and state space of a whole-body robot. IL, on the other hand, usually suffers from the distribution shift problem due to the lack of high-quality demonstrations over numerous scenarios\cite{fang_survey_2019}. Hence, it remains very challenging for high-DoF robots to learn complex manipulation skills and further generalize these skills to novel objects.

\begin{figure}
\vspace{0.4em}
\setlength{\abovecaptionskip}{0.cm}
\centering
\includegraphics[width=1\linewidth]{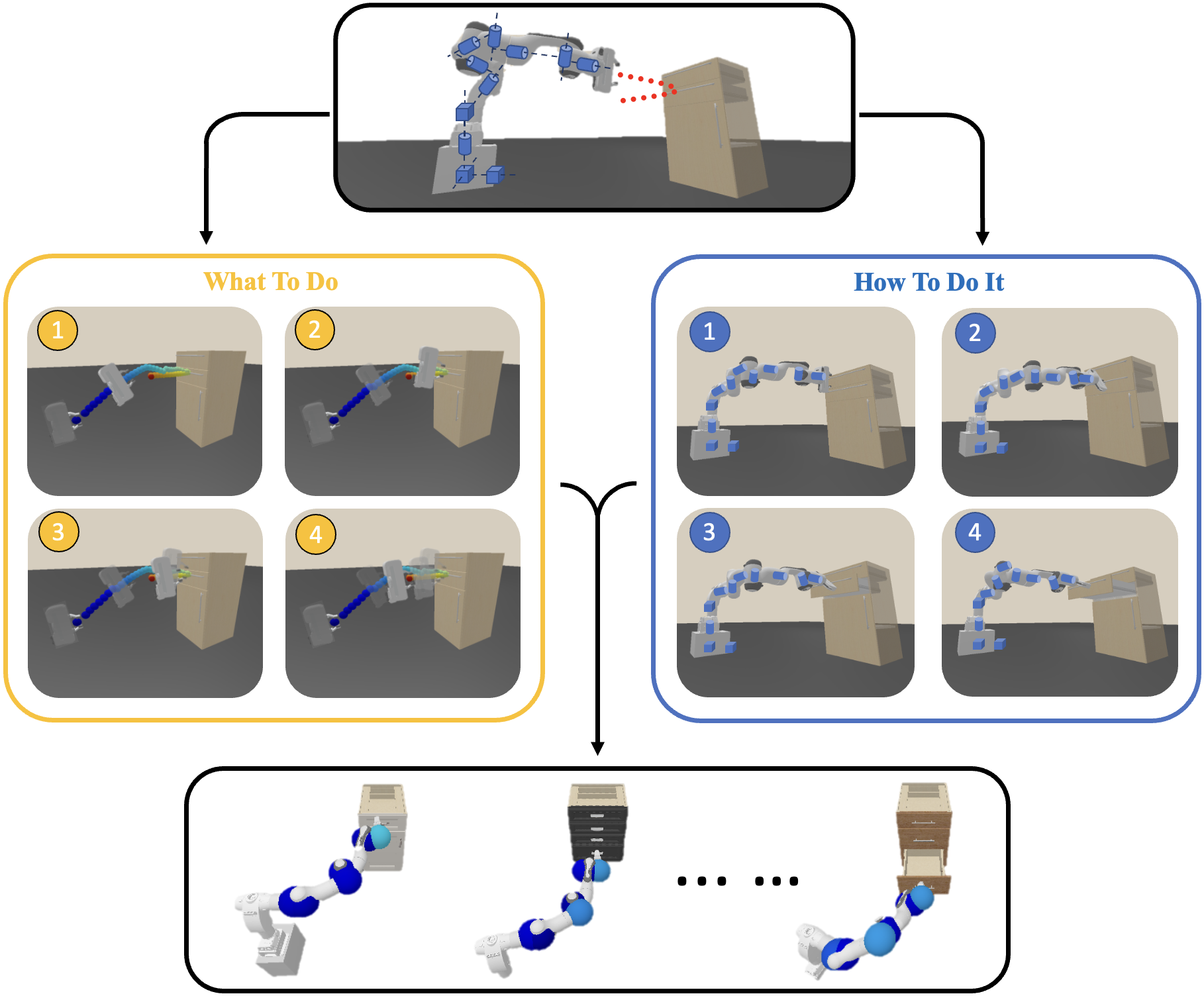}
\caption{\textbf{Demonstration of decoupling skill dynamics.} The left panel shows how we use reinforcement learning to determine how to control a disembodied or floating end-effector (`what to do'). The right panel shows how we use quadratic programming to realize a compliant trajectory and set of joint-torque control signals (`how to do it'). The bottom panel shows how we can use this to generalize to different, previously unseen scenarios. }
\label{fig:core}
\vspace{-1.5em}
\end{figure}

We note that humans and other animals can easily generalize a learned skill from one object to another, and execute the skills even if a limb is injured or being constrained. We take a preliminary step towards providing robots with similar capabilities. To achieve this, we propose that the complex yet low-DoF \textit{skill dynamics} (i.e., what should be done) can be decoupled from robot-specific kinematic control (i.e., how this behavior needs to be implemented). The skill dynamics themselves can be learned from manipulator-object interactions, without being concerned with high-DoF whole-body control. Whole-body control can be treated as a classical inverse kinematics problem. We conjecture that the robot could better learn generalizable manipulation skills through this decoupling, by making the RL tasks simpler.

The most inspired works to ours are the action-primitive methods. Dalal et al. \cite{dalal_accelerating_2021} manually specify a library of robot action primitives such as `lift' and then learn an RL policy to parameterize these primitives such as `lift to position (x,y)'. Alternatively, Pertsch et al.  \cite{pertsch_accelerating_2020} propose to learn a prior over basic skills from offline agent experience, which allows the robot to explore these skills with unequal probabilities. These methods decompose the agent's action space into high-level policy and low-level predefined behaviors, leading to improvement in learning efficiency and task performance. However, a potential concern is that not all manipulation skills can be represented as a series of action primitive\cite{dalal_accelerating_2021}. In this paper, we do not expect that a robot needs to learn manipulation skills by manually creating action libraries.

To realize our approach, we learn the skill dynamics with a `disembodied' manipulator (effectively a free-floating end-effector) via model-free RL and then control the actual robot via QP optimization based on its full physical model. The general idea is appealing since learning skill dynamics from interactions enhances the robot's understanding of its surroundings and objects, leading to diverse manipulation trajectories and generalizability to differing objects. With regard to the RL agent, controlling a disembodied manipulator greatly reduces the dimensionality and complexity of the search space. 
In our approach, the whole-body controller formulates a QP model with robotic singularity and kinematic constraints. 
The high-dimensional joint-space actions are produced by: 1) predicting end-effector (EE) poses and velocities from learned skill dynamics (i.e., the RL agent), 2) optimizing the joint velocities by a QP solver, 3) automatically generating joint torques by inner PID controllers of the simulator. Our method thus accelerates the RL process and produces more compliant, smoother, and controllable robot motions. In converse, using pure RL for controlling whole robots can generate jerky or `locked' motions due to singularities, leading to a high task failure rate.

We evaluate this approach using the ManiSkill  robotic simulated environment \cite{mu_maniskill_2021} consisting of a 13-DoF Franka robot, onboard RGB-D cameras for point cloud observations, and a variety of articulated objects.
Experiments show that our approach can learn a generalizable skill policy over different cabinets of the training set and the unseen test set. We achieve an average success rate of 74\% on training cabinets and a 51\% on test cabinets for the drawer opening task, 
outperforming existing techniques in \cite{mu_maniskill_2021}. 
We show that the generalization of our model is improved by increasing the number of training objects.  We also compare the robotic motions produced by our method and pure RL, showing that robotic singularities can be avoided by our method. 

\section{RELATED WORK}
Robot learning for manipulation skills has been intensively studied in recent years. The dominant methods are reinforcement learning and imitation learning. 
An extensive survey of existing methods can be found in Kroemer's work \cite{kroemerreview}. 
However, robotic manipulation of complex objects through controlling high-DoF robots remains a challenging problem.

Recent works have applied a variety of RL algorithms for robotic manipulation \cite{gu_deep_2017, kalashnikov_qt-opt_2018, yamada_motion_2020, liu_distilling_2021}. However, productive exploration has always been a challenge for robotic RL due to the complex dynamics of high-DoF robots. To address this, many works have looked into decomposing the action space into higher-level policy and lower-level control. Exploiting predefined action primitives (e.g., lift, move, slide) is a convenient way to reduce the action space dimensions\cite{dalal_accelerating_2021, pertsch_accelerating_2020, xia2021relmogen}, 
but it usually requires manual specification of robotic behaviors.
Another category of approaches looks into operational space control (OSC) \cite{khatib1987osc, peters2007reinforcement, wong2022oscar}, which maps the desired EE motions into joint torques but requires a precise mass matrix. Martin et al. \cite{martin-martin2019variable} compare OSC and other control modes of EE space and joint space in the Robosuite simulated environment\cite{zhurobosuite}. While OSC-based RL methods usually perform better than other control modes in contact-rich tasks\cite{martin-martin2019variable}, they are highly sensitive to inaccuracies in dynamics modeling\cite{nakanishi2008operational,wong2022oscar}.
Our method shares a similar motivation with OSC-based RL, but our agent is a disembodied manipulator rather than a complete robot, without the need for high-fidelity modeling of the mass matrix during RL training. Furthermore, our work presents category-level generalizability across various objects in the ManiSkill environment, while tasks in Robosuite typically focus on a single scene.

\begin{figure}
\setlength{\abovecaptionskip}{-0.2em}
\centering
\includegraphics[width=1.0\linewidth]{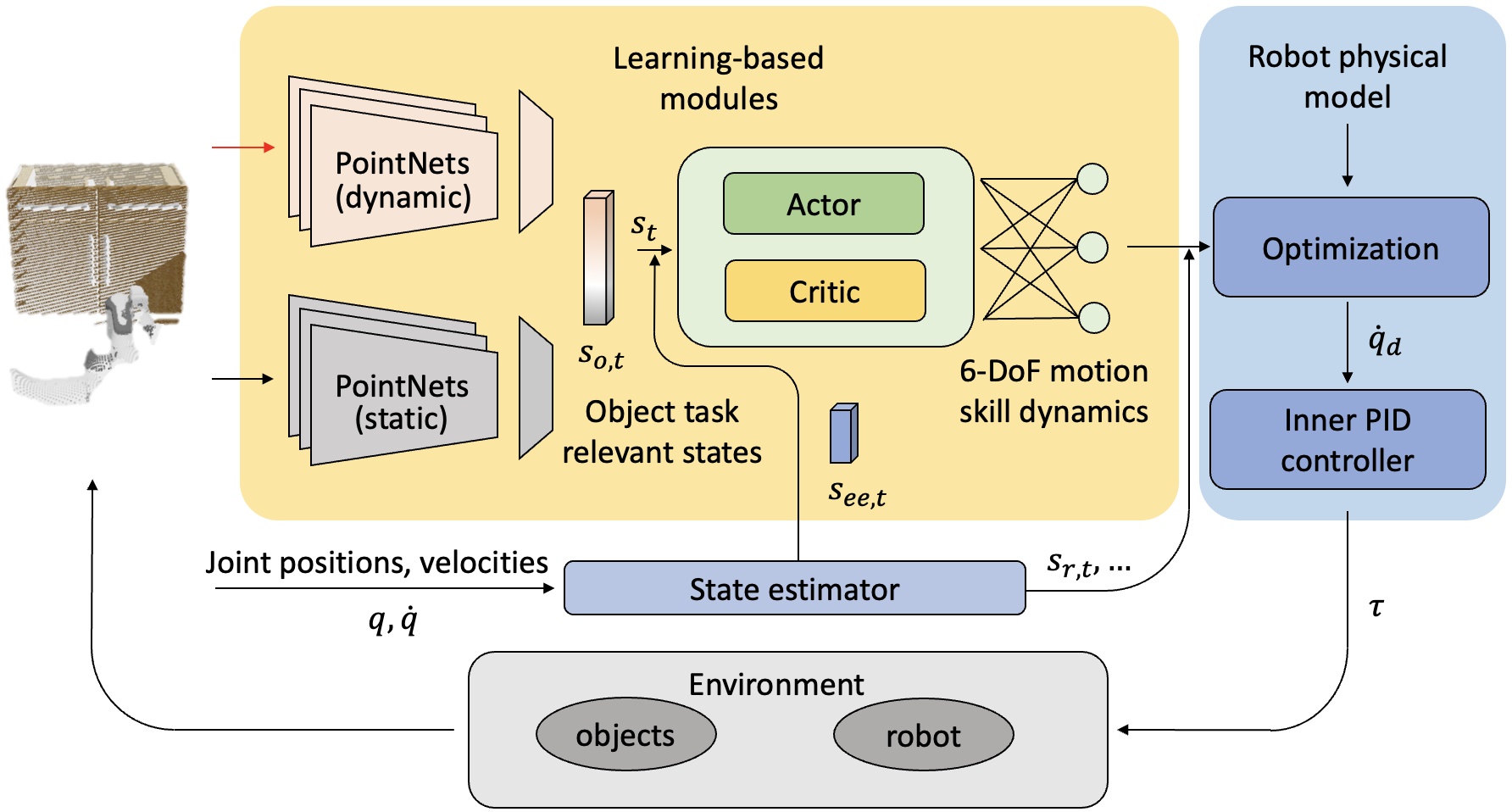}
\caption{\textbf{Architecture of the proposed system.} As shown in the yellow block, we use two simple PointNet\cite{qi_pointnet_2017} ensembles to separately perceive static (e.g., size of an object) and dynamic (e.g., current position of a handle) states. These are inputs to a SAC RL framework to learn how to control the disembodied end-effector, realizing a 6-DoF motion skill. Through knowledge of the robot's physical model, QP is used to optimize control of the joint dynamics of the whole-body robot.}
\label{fig:pipeline}
\vspace{-1.4em}
\end{figure}

Imitation learning has usually been considered a more sampling-efficient approach, but collecting high-quality demonstrations that can be directly applied to the robot motors is usually difficult \cite{fang_survey_2019}. Shen et al. \cite{shen2022gail} propose a generative adversarial self-imitation learning method and several techniques such as instance balancing expert buffer to mitigate the issues in ManiSkill's demonstration trajectories. Imitation learning for manipulation policies using these techniques has shown impressive improvements, but this work focuses more on learning from interactions.

Another highly related topic is robotic vision for articulated object manipulation. Mo et al. \cite{mo_where2act_2021} and Wu et al. \cite{wu_vat-mart_2021} utilize encoder-decoder networks to extract affordance maps and predict actions or trajectory proposals attached to the 3D object representations. 
Eisner et al. \cite{eisner2022flowbot3d} also develop a neural network to predict potential point-wise motions of the articulated objects from point cloud observations. Similarly, Xu et al. \cite{xu2022umpnet} suggest predicting potential moving directions and distances of the articulated parts but using a single image as input.  Alternatively, Mittal et al. \cite{mittal2022articulated} explicitly infer the articulation properties such as joint states and part-wise segmentation by giving object categories. 
These methods are object-centric representation learning and more focus on object articulation properties, usually leaving the next manipulation stage in a simplified way (e.g., a perfect suction cup in \cite{xu2022umpnet}), while our work learns an agent policy for a disembodied hand to generate continuous and close-loop actions and then performs whole-body control for the robot.

\section{Reinforcement Learning for Skill Dynamics}
\label{sec: rl}

Fig. \ref{fig:pipeline} shows the overall framework of our approach. We first learn the control policy of a disembodied manipulator as shown in the yellow block, and then compute the whole robot's high-dimensional joint actions by QP optimization as illustrated in the light blue block. 
We shall discuss the RL policy and visual perception in the current section, leaving the discussion of the whole-body controller in section \ref{sec:qp}. 


\subsection{Problem Formulation}
Robotic manipulation skill learning is usually formulated as a Markov decision process (MDP), which is represented as $(S,A,R,T,\gamma)$, where $S$  is the set of states, $A$ is the set of actions, $R(s_t,a_t,s_{t+1})$ is the reward function, $T(s_{t+1}|s_t,a_t)$ is the transition function as a probability distribution,  and $\gamma$ is the discount factor for the future rewards. The agent policy $\pi(a|s)$ represents the action selecting probability under a given state $s$.  The goal of RL is to maximize the return under the policy $G_\pi = \mathbb{E}_\pi [\sum_t \gamma^{\; t} R(s_t,a_t,s_{t+1})]$. In vision-based RL, we usually need to estimate the task-relevant states from observation $O$. This setting is viewed as a partially observable Markov decision process (POMDP), and the policy is regarded as $\pi (a|f(o))$ where $f(o)$ is the estimated states or features from visual representations. 

\subsection{Disembodied Manipulator Environment}

\begin{figure}
\setlength{\abovecaptionskip}{0.cm}
\centering
\includegraphics[width=1\linewidth]{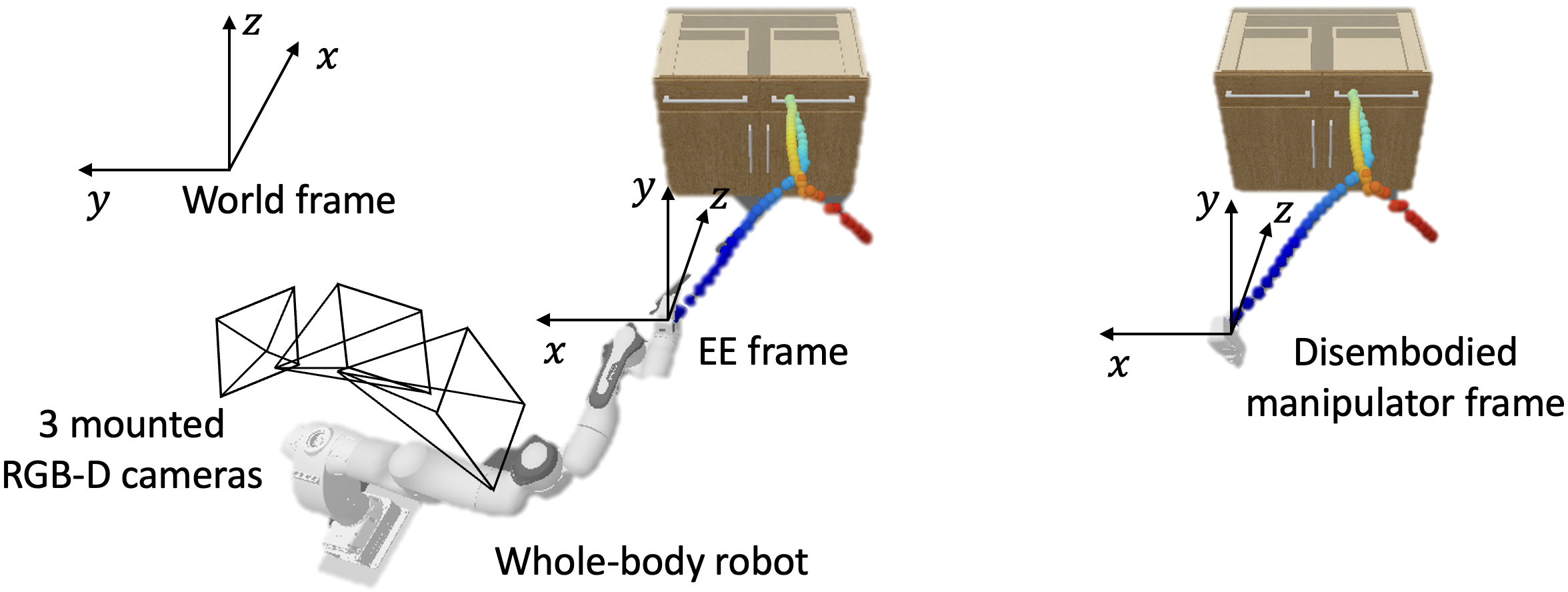}
\caption{\textbf{Coordinate system of our method.} During the manipulation process, we optimize the joint-space actions of the robot to approximate its end-effector (EE) motions to the disembodied manipulator's trajectory. At every time step, the ego-centric point cloud observation is obtained from the three RGB-D cameras mounted on the robot.} 
\label{fig: coord}
\vspace{-1em}
\end{figure}

The agent for learning skill dynamics is a 6-DoF disembodied (or floating/ flying) manipulator with two fingers (as shown in Fig. \ref{fig: coord}). The DoF of the manipulator is realized by three virtual prismatic joints and three virtual revolute joints. Therefore, the action space is six desired velocities of the virtual joints $v_{h,d} \in  \mathbb{R}^6$ and two desired positions of the finger joints $q_{f,d} \in  \mathbb{R}^2$. 
Following the settings in ManiSkill, the action space is normalized to $(-1,1)$ in RL training:
\begin{equation}
a = [v_{h,d}, q_{f,d}]
\end{equation}

The state space of a manipulation task contains the manipulator states $s_{ee}$ and the target object states $s_o$: 
\begin{equation}
s =[s_o, s_{ee}]
\end{equation}
For the cabinet environment of ManiSkill, we define $s_o=[s_{cab}, s_{link}, s_{hdl}, s_{size}]$, where $s_{cab}$ is the base link pose of the loaded cabinet (note that the cabinet base is currently fixed to the ground following the environment default setting), $s_{link}$ and $s_{hdl}$ are the current poses and the full poses (i.e., the poses when the drawer is fully opened) of the target drawer link and the handle, and $s_{size}$ is the full length and the opening length of the target drawer. The poses are all represented as world frame coordinates and quaternions. And $s_{ee} = [s_{vq}, s_{f}]$ is the agent states, where $s_{vq}$ is the positions and velocities of the disembodied manipulator's virtual joints, and $s_f$ is the world frame positions and velocities of the two fingers. 

We follow the original design of the dense reward from ManiSkill which is well-shaped and MPC-verified\cite{mu_maniskill_2021}.
For the cabinet drawer opening task, the reward is defined as:
\begin{equation}
    R = 
\begin{cases}
R_{stg} + R_{ee}, \; d>0.01 \\
R_{stg} + R_{ee} + R_{link}, \; d<0.01, \; c_{op} < 0.9 \\
R_{stg} + R_{ee} + R_{link} + R_{stc}, \; d<0.01, \; c_{op} > 0.9
\end{cases}
\end{equation}
where $R_{stg}$ increases from the first stage to the final and the stage is defined by the distance between EE and the handle of the target drawer $d \in \mathbb{R}$ and the opening extent of the target drawer $c_{op} \in [0,1]$. More detailed definitions and coefficients can be found in \cite{mu_maniskill_2021}. The reward $R_{ee}$ encourages the fingers to get closer to the handle, $R_{link}$ encourages the target drawer link to be manipulated to its goal, and $R_{stc}$ expects the drawer to be static after task completion. 




\subsection{Soft Actor-Critic Algorithm for Manipulation}
In order to learn the skill dynamics, we adopt the model-free soft actor-critic (SAC)\cite{haarnoja2018sac} 
algorithm for policy learning, where a Gaussian policy head is used for the continuous action space. The actor and critic functions are approximated by multi-layer perceptions (MLPs). The parameters of the MLPs are updated as off-policy RL with the samples from a replay buffer $D_{replay}=\{(s_t, a_t, r_t, s_{t+1}) | t=0,1,2...\}$. 

\begin{figure}
    \centering
    \subfloat[Learning curves of RL models.
    ]
     {
         \hspace{-5.mm}\includegraphics[width=0.9\linewidth]{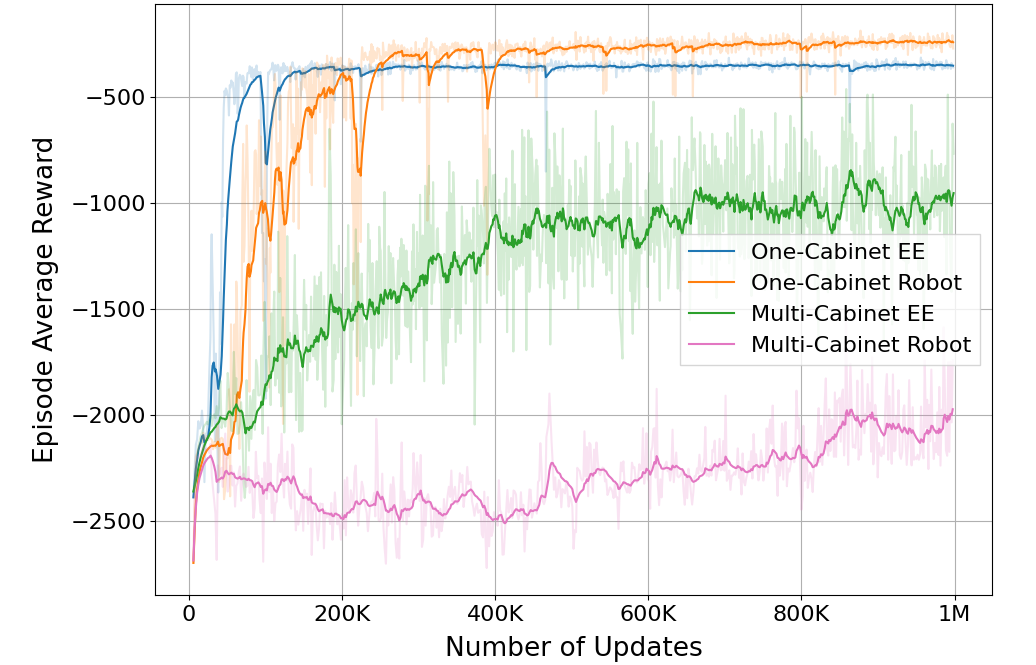}
         \label{fig:train_curve}
     }
    \vspace{-1mm}
     \subfloat[Evaluation on training cabinets.
     ]
     {
         \hspace{1.3mm}\includegraphics[width=0.92\linewidth]{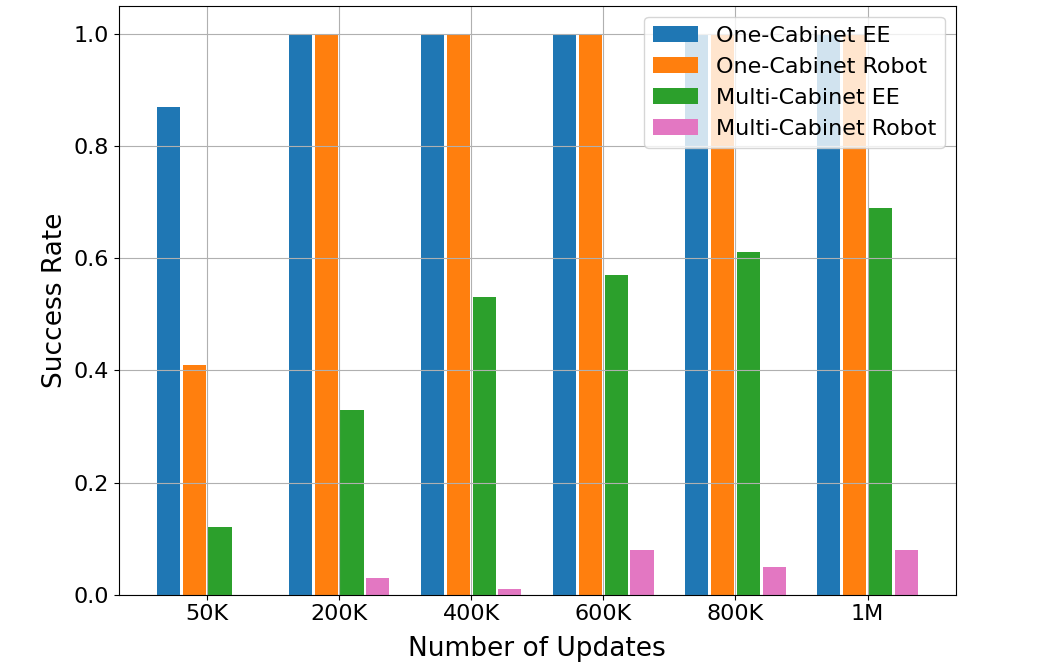}
         \label{fig:train_suc}
     }
     \caption{\textbf{Comparing RL agents of the disembodied manipulator and the whole-body robot.} We compare the agents under two conditions: training with a single cabinet or multiple (i.e., 15) different cabinets. 
     While both agents succeed in the single-cabinet environment, the disembodied manipulator notably outperforms the robot in the multiple-cabinet environment.}
    \hspace{1px}
    \label{fig:hist}
\vspace{-1em}
\end{figure}

The learning curves of our RL agent of the disembodied manipulator and the original SAC agent of the whole-body robot by ManiSkill-Learn \cite{mu_maniskill_2021} are shown in Fig. \ref{fig:train_curve}. 
For an identical cabinet, both the robot and the manipulator successfully learn the specific skill, but the manipulator learns more quickly. We notice that the whole-body robot finally obtains a higher episode average reward than the manipulator, because the robot's EE can move much faster when the robot does not fall into kinematic singularities, resulting in a shorter episode length of task completion. However, the evaluation results on multiple different cabinets in Fig. \ref{fig:train_suc} show that the whole-body robot with a high-dimensional action space hardly succeeds to learn the required manipulation skill, while our manipulator can still learn the skill dynamics over a variety of objects with high task success rates.



\subsection{3D Visual Perception}

The vision module of our work is regarded as a state estimator, mapping the 3D observations to states $s_o$. Similar to the baselines in \cite{mu_maniskill_2021}, we use PointNet~\cite{qi_pointnet_2017} as the backbone which takes masked point clouds as input. To achieve a more stable state estimation, we adopt two separate ensembles of PointNets for estimating static states (e.g., the size of drawers) and dynamic states (e.g., the current pose of the handle) respectively. In this work, we use single-frame prediction and fully-supervised learning for the vision module. We leave the unsupervised visual representation learning from interactions (LfI)\cite{xia_interactive_2020, zeng_learning_2018,pinto2016supersizing} for future exploration.

The pairs of point clouds and states for training are collected by synchronizing the robot's EE and the disembodied manipulator via inverse kinematics. Note that the point clouds are partial observations of the cabinets, as they are generated from the data captured by the three cameras mounted on the robot. We also synchronize the cabinet states in both the robot's and the manipulator's environments, then we collect observations from the robot. The training set can be represented as $D_{vis}=\{(o_{pcd,t}, s_{o,t}) | t=0,1,2...\}$, where $o_{pcd,t}$ are 1200 points observed at time step $t$ with colors and point-level task-relevant masks (i.e., handle of the target drawer, target drawer, and robot) following the ManiSkill settings. 
Finally, the PointNets are followed by MLPs to predict $\hat{s}_{o,t}$, and the vision module is trained with an L2 loss between the ground truth $s_{o,t}$ and the predicted $\hat{s}_{o,t}$. 



\section{QP-Based Whole-Body Controller}
\label{sec:qp}


The second phase of our framework is to calculate the joint-space actions for whole-body control based on the robot's physical model. Our goal is to control the robot's EE to execute the learned skill dynamics in the same way as the disembodied manipulator. The poses and velocities of the manipulator are accurately calculated via forward kinematics (FK), and then we synchronize the robot's EE to approximate the trajectory which can be viewed as an inverse kinematics (IK) problem. However, the robot in our environment consists of a movable base and the off-the-shelf IK solver usually returns solutions such that the EE displacement is mainly contributed by the base joints.  Moving the base instead of the arm is usually slower and more energy-consuming. To decrease the base motion while avoiding robotic singularities, we introduce QP optimization to calculate the desired joint velocities for controlling the EE's motion. In this work, we use Klampt\cite{Klampt} for robotic kinematics calculations and qpOASES\cite{Ferreau2014} as the QP solver.

According to forward kinematics, the robot's EE pose $x_{ee} \in \mathbb{R}^m$ can be represented by the robot's current configuration $q$ (i.e., joint positions):
\begin{equation}
x_{ee} = p(q), q \in \mathbb{R}^n
\end{equation}

The Jacobian matrix of EE in the current configuration indicates the influences of each joint's motion to EE: 
\begin{equation}
\label{eq:jacobian}
J(q)=
\begin{bmatrix}
    \partial p_1 / \partial q_1 & \dots & \partial p_1 / \partial q_{n} \\
    \vdots & \ddots & \vdots \\
  \partial p_m / \partial q_1  & \dots & \partial p_m / \partial q_n
\end{bmatrix}
\end{equation}
where $m$ is the DoF of the robot's workspace and $n$ is the DoF of the joint space. Then the velocity is:
\begin{equation}
\dot{x}_{ee} 
=J(q) \dot{q} 
\end{equation}

The task of approximating the EE motion to the disembodied manipulator can be written as:
\begin{equation}
T_{ee} = J(q) \dot{q} - \dot{x}_{ee,d}
\end{equation}
where $\dot{x}_{ee,d}$ is the EE's desired velocity. As the whole-body robot's EE is represented as the Panda hand in our setting, the fingers will not be considered in the QP optimization. The finger actions will directly follow the learned RL policy. Thus, we can regard $q$ as the 11-DoF of the robot in this section, excluding the 2-DoF of the fingers. At time step $t$, we use the trajectory of the disembodied manipulator $\{x_{h,t}\}$ to obtain the desired pose change of EE:
\begin{equation}
    \dot{x}_{ee,d}= \frac{\triangle x_{ee,d}}{\triangle t } = \frac{(x_{h,t+\triangle t} - x_{ee,t})}{\triangle t }  
\end{equation}
The reason for not directly using the predicted action $v_{h,d}$  from the RL policy to compute $\dot{x}_{ee,d}$   is that $v_{h,d}$ represents the target joint velocity commands for the manipulator's virtual joints so it does not indicate the actual desired pose change.

Robotic singularity refers to degenerate joint configurations that cause the rank of the Jacobian matrix to reduce. If $rank(J_{ee})$ drops below the workspace dimension, the robot cannot move instantaneously in a certain workspace dimension. 
It is dangerous when robots are in singular or near-singular configurations, as these usually cause extreme joint accelerations. To avoid these conditions, we expect that the robot should be able to move the EE flexibly in the current configuration. 
We attempt to parameterize this by imposing kinematic constraints on the robot's workspace, regarded as the robot's capability of moving the EE to the desired pose after $k$ steps while the base is fixed, using the weight matrix $C_{sg}$ of the QP's cost function:
\begin{equation}
    C_{sg} = [c_{base}I_{base}, \; c_{arm}I_{arm}] \in \mathbb{R}^{n \times n }
\end{equation}
where $c_{base}, c_{arm}$ are the weight coefficient and $I_{base}, I_{arm}$ are the identity matrix. We use the learned RL policy to predict the desired EE's pose after $k$ steps as $\hat{x}_{ee, t+k \triangle t} \ $ and then perform IK via Klampt. We obtain
$z_{res}, \hat{q}_{t+k \triangle t} = f_{ik}(q_{t}, \hat{x}_{ee, t+k \triangle t})$
where $z_{res}=1, \hat{q}_{t+k \triangle t} \neq \emptyset$ when IK is solved, and $z_{res}=0, \hat{q}_{t+k \triangle t} = \emptyset$ when the numerical solver fails to find a feasible configuration, which indicates that EE is unable to reach the goal pose $\hat{x}_{ee, t+k \triangle t}$ after $k$ steps if the base remains static during this period. Thus, we can define:
\begin{equation}
\begin{gathered}
c_{base} = \omega_1 z_{res} + \omega_2 (1-z_{res}) \\c_{arm} = \omega_2 z_{res} + \omega_1 (1-z_{res})
\end{gathered}
\end{equation}
QP is then applied to optimize desired joint velocities $\dot{q}_{d}$:
\begin{equation}
\begin{aligned}
    \underset{\dot{q}_{d}}{\operatorname{minimize}}  \quad &\dot{q}_{d}^T C_{sg} \dot{q}_{d} \quad \quad \\
\text {subject to \ } \;
&J(q) \dot{q}_{d} =\dot{x}_{ee,d}  \\ 
&\dot{q}_{\min } \leq \dot{q}_{d} \leq \dot{q}_{\max } 
\end{aligned}
\end{equation}
where $\omega_1 > \omega_2 $, the QP tends to encourage base motion when the robot's EE cannot move to $\hat{x}_{ee, t+ k \triangle t}$ with its base fixed (i.e., $c_{arm}=\omega_1, c_{base}=\omega_2$ under this condition), which accordingly avoids robotic singularities in most cases. In our experiment, we set $k=10, \omega_1=20, \omega_2=0.1$ for the QP.

\section{Experiments}
\label{sec:exp}

In this section, we evaluate our method in three aspects: model improvements during interactions, policy generalizability to novel objects compared to baselines, and motion compliance of the whole-body robot.

The experiments are performed in the ManiSkill simulated environment \cite{mu_maniskill_2021} built on SAPIEN \cite{xiang_sapien_2020}. 
In particular, we apply our approach to the cabinet drawer opening task, where the success condition is opening the target drawer's joint to 90\% of its extent. The task is episodic with a time limit of 200 steps. The agents for manipulation are the disembodied manipulator and the default single-arm robot that consists of a movable dummy base, a Franka arm, and a Panda gripper. The initial pose of agents and the selection of cabinets are randomized by seeds at the start of every episode.

Based on the public-available 25 cabinets verified for the drawer opening task, we randomly split them into a training set of 15 cabinets and a test set of 10 cabinets. The agent will not interact with or see the test cabinets during training.


\begin{figure}
\setlength{\abovecaptionskip}{-2mm}
\centering
\includegraphics[width=0.97\linewidth]{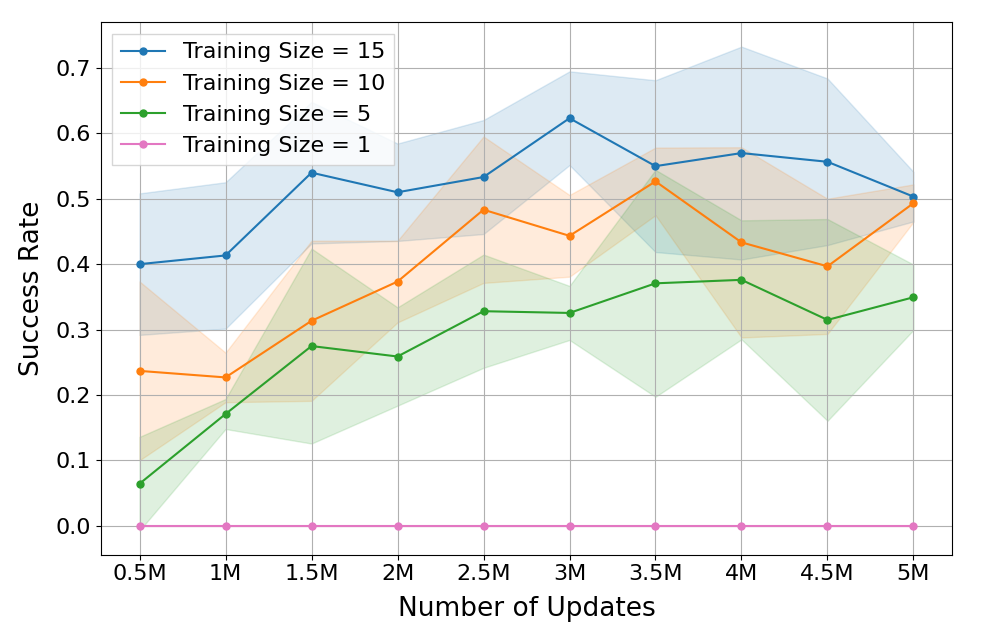}
\caption{\textbf{Average success rates on unseen cabinets using RL models with varying training set sizes.} We show a 95\% confidence interval of the mean performance on test cabinets across three random seeds (100 episodes per seed). The training set size is the number of cabinets for training.}
\label{fig:diff_size}
\vspace{0em}
\end{figure}

\begin{figure}
    \centering

    \subfloat[Failure, step = 200, training size = 5.]
     {
         \includegraphics[width=0.14\textwidth]{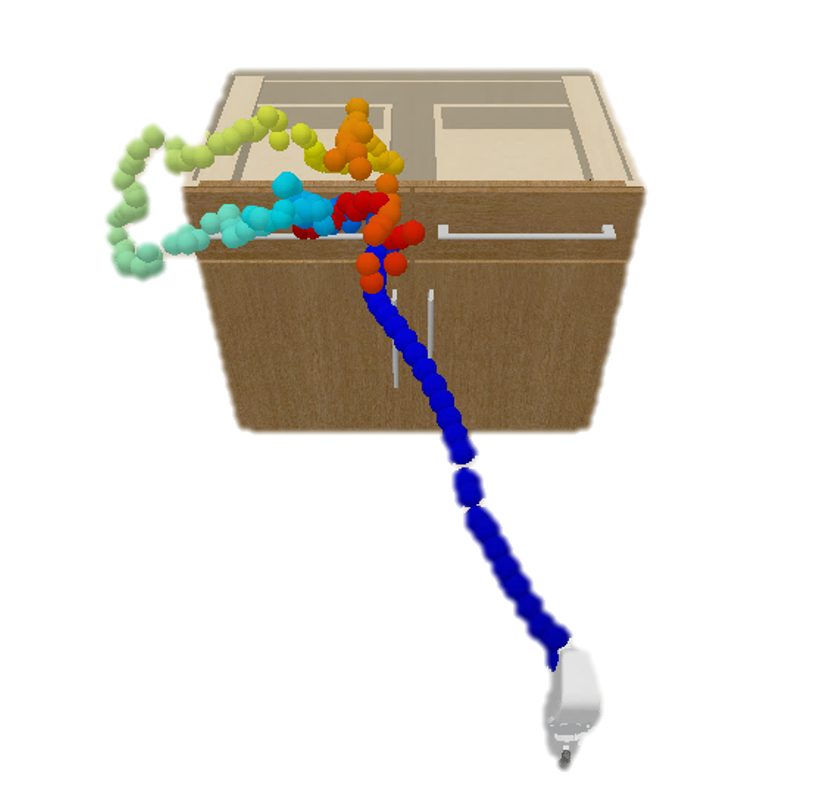}
         \label{fig:vt1}
     }
     \hspace{0mm}
     \subfloat[Success, step = 164, training size = 10.]
     {
         \includegraphics[width=0.14\textwidth]{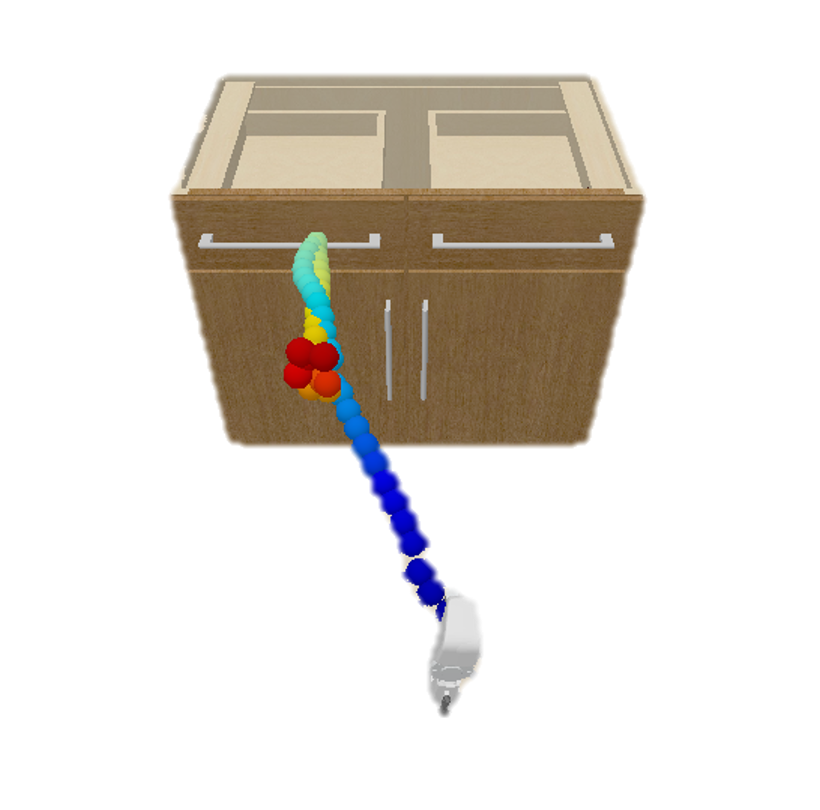}
         \label{fig:vt2}
     }
     \hspace{0mm}
     \subfloat[Success, step = 50, training size = 15.]
     {
         \includegraphics[width=0.14\textwidth]{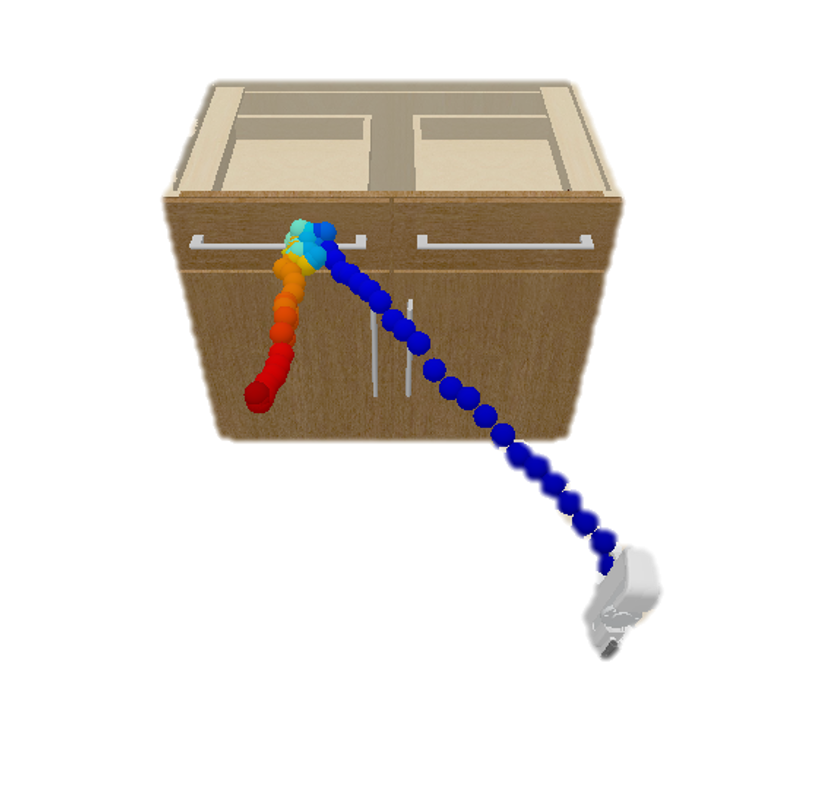}
         \label{fig:vt3}
     }
     \caption{\textbf{Visualization of trajectories on an unseen cabinet using RL models with varying training set sizes.} By interacting with more cabinets, the RL model shows a better understanding of the skill dynamics, resulting in smoother and more reasonable motions.}
    \hspace{0px}
    \label{fig:vis_traj}
\vspace{-1em}
\end{figure}

\subsection{Model Performance with Varying Training Set Sizes}

We demonstrate our RL models improving during interactions with more cabinets in the environment (as shown in Fig. \ref{fig:diff_size}) by varying training set sizes (i.e., the number of different cabinets for training). We notice that RL training converges faster with smaller training sets, but such models lack generalizability to novel cabinets and the worst case can be completely unsuccessful. With more cabinets seen in training, the policy model achieves higher task success rates on the test set. The quantitative results in Fig. \ref{fig:diff_size} indicate a continuously evolving process of the RL model learning a generalizable manipulation skill policy. 

Fig. \ref{fig:vis_traj} shows the qualitative results by visualizing the trajectories of the disembodied hand on a test cabinet. The policy models for generating these trajectories are trained with different numbers of cabinets. We observe that models with fewer training cabinets show worse skill dynamics. For example, the RL agent trained with five cabinets can move forward in the correct direction to the target link, but it either produces confusing motions near the handle or pulls the drawer back and forth, while the other models present more reasonable motions and shorter episode lengths. 


    


\begin {table}[!ht]
\setlength{\abovecaptionskip}{0cm}
\setlength{\belowcaptionskip}{-0.1cm}
\centering
\caption{Task success rates of different approaches}
\begin{tabular}{m{1.5cm}<{\centering}|m{1.5cm}<{\centering} m{1cm}<{\centering}|m{1.5cm}<{\centering} m{1 cm}<{\centering} }
\hline
Metric & \multicolumn{2}{c|}{Average Success Rate} & \multicolumn{2}{c}{Episode Average Length} \\
\hline
Split & Train & Test & Train & Test  \\
\hline
BC-robot  & 0.38 & 0.13 & 137 & 178 \\

BCQ-robot  & 0.25 & 0.11 & 166 & 184 \\

SAC-robot  & 0.08 & 0.03 & 189 & 195 \\

Ours-ee-s & 0.86 & 0.63 & 77 & 101 \\

Ours-robot-s & 0.83 & 0.61 & 90 & 114  \\

Ours-ee-v & 0.78 & 0.55 & 93 & 119\\

Ours-robot-v  & 0.74 & 0.51 & 95 & 129\\

    \hline
\end{tabular}
\label{tab:simulation_result}
\vspace{-1em}
\end{table}

\subsection{Generalizability Comparison with ManiSkill Baselines}

We compare our method with the original RL and IL baselines in ManiSkill \cite{mu_maniskill_2021} via two metrics: Average Success Rate and Episode Average Length. The average success rate is calculated from three runnings on the test set, where each running uses a different random seed to initialize 100 episodes, and all the models are trained to 3M steps.

Our baselines include Behavior Cloning (BC), Batch-Constrained Q-Learning (BCQ), and SAC for whole robot control (SAC-robot). These methods directly predict the joint-space actions to control robots, where BC and BCQ utilize a PointNet+Transformer architecture and SAC uses MLPs. We compare different combinations of modules of our method, where `ee' and `robot' represent the disembodied manipulator or the full robot, `s' and `v' represent using states or point clouds as input, respectively.  

Table \ref{tab:simulation_result} shows that our method significantly outperforms the baselines, indicating better generalizability over novel objects. Our state-based disembodied manipulator achieves the best performance on training and test cabinets, which indicates an upper bound of the proposed framework. 
As for our vision-based agents, they show slightly lower performance by adding the whole-body controller and the vision module. However, the overall results still surpass the baselines. 
These demonstrate that the decoupling approach is more suitable for generalizable skill learning than the baselines.

\subsection{Visualization of Robot Motion Compliance}

\begin{figure}
\setlength{\abovecaptionskip}{0.cm}
\hspace{2mm}\includegraphics[width=1\linewidth]{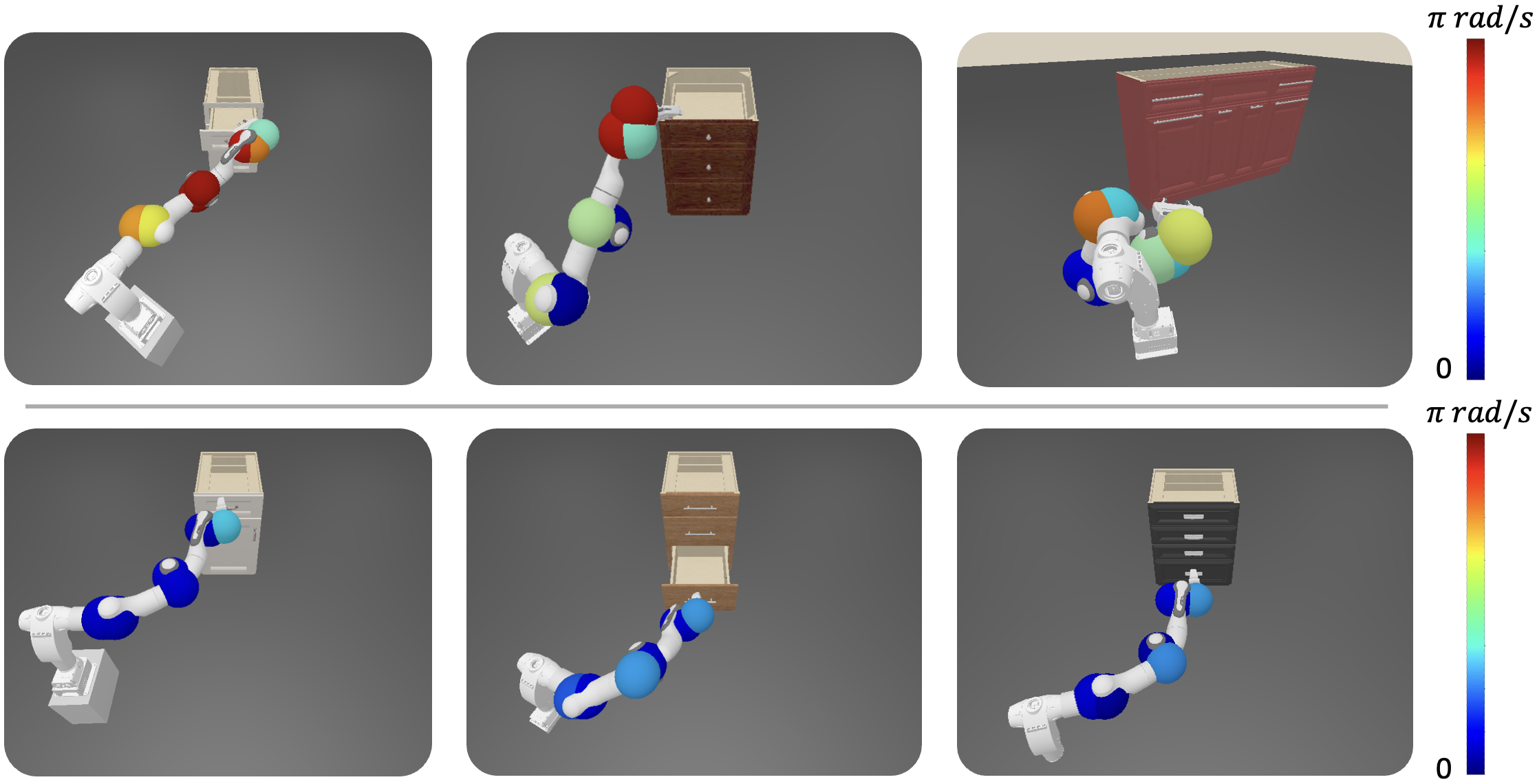}
     \caption{
     \textbf{Typical robot motions generated by BCQ (top) and our method (bottom), colored by joint velocities. Note how BCQ exhibits far higher joint velocities.} The BCQ method learns a  direct policy to predict the desired joint velocities for robot control, while our method decouples the policy learning and controls the whole-body joints by QP optimization.}
    \label{fig:singular}
\vspace{0em}
\end{figure}

In addition to category-level generalization to objects, our approach naturally produces compliant and controllable robot motions. It is essential for high-DoF collaborative robots to ensure motion compliance, interpretability, and safety.
Fig. \ref{fig:singular} shows exemplar traces of robots in singular configurations manipulating objects. Due to the decreasing rank of the EE's Jacobian matrix, the robot joints are required to accelerate to an extremely high velocity to perform the workspace behaviors. The red balls show the correlated joint velocities hit the limits ($\pi \; rad/s$ in our case). A reason for this problem is that high-dimensional actions are challenging to learn by RL e.g., robots in Fig. \ref{fig:singular} (top) are controlled by a BCQ policy. Therefore, the policy may tend to reduce the whole-body DoF causing `stretching' or `curling' motions.

Our approach takes advantage of the robotic physical model and the established optimization solver. The whole-body actions are well-controlled while the EE follows the learned policy of the disembodied manipulator. Fig. \ref{fig:singular} (bottom) shows that our robot applies much lower joint velocities than BCQ but completes the task with higher success rates. 


\subsection{Ablation Study on Random Splits}

We perform an ablation study on random splits of training and test set. Table \ref{tab:ablation_results} shows the average success rate and episode average length of the disembodied manipulator model on three different splits. The overall performances are similar, which demonstrates that our approach is robust to the randomly splitting manner. 



\begin {table}[!ht]
\setlength{\abovecaptionskip}{0cm}
\setlength{\belowcaptionskip}{-0.1cm}
\centering
\caption{Task success rates with random training and test splits}
\begin{tabular}{m{1.5cm}<{\centering}|m{1.5cm}<{\centering} m{1cm}<{\centering}|m{1.5cm}<{\centering} m{1 cm}<{\centering} }
\hline
Metric & \multicolumn{2}{c|}{Average Success Rate} & \multicolumn{2}{c}{Episode Average Length} \\
\hline
Split & Train & Test & Train & Test  \\
\hline
Set 1-s & 0.93 & 0.59 & 66 & 124\\
Set 2-s & 0.86 & 0.63 & 77 & 101 \\
Set 3-s & 0.87 & 0.60 & 74 & 115\\
Set 1-v & 0.82 & 0.54 & 85 & 118\\
Set 2-v & 0.78 & 0.55 & 93 & 119 \\
Set 3-v & 0.79 & 0.52 & 88 & 125\\
\hline
\end{tabular}
\label{tab:ablation_results}
\vspace{-2em}
\end{table}

\subsection{Limitations and Future Work}
While we present that decoupling policy learning from robot-specific control is effective for generalizable object manipulation, there are several limitations we need to address in future work. One concern is that the evaluation is limited to a single drawer opening task and pure RL and IL baselines. As robot learning for manipulating complex 3D objects is a highly challenging area, we choose this representative task i.e., opening a drawer or cupboard as a starting point, and we notice that there are many concurrent works\cite{dalal_accelerating_2021,xu2022umpnet,mo_where2act_2021, wu_vat-mart_2021,eisner2022flowbot3d,shen2022gail} presented. We will incorporate these methods as baselines and evaluate more tasks in the future. Another potential issue is that sim-to-real transfer usually requires non-trivial efforts. In our approach, since the robot learns the skill policy by interacting with various objects, we believe it would be robust when facing the domain gap. Furthermore, we would like to explore both object-level and robot-level generalizability, i.e., generalizing the learned policy to robots with joint injured or being constrained, in our future work.

\section{conclusion}
The paper proposed a manipulation skill learning method for high-DoF robots where the skill dynamics are decoupled from robot-specific kinematic control. In the first stage, a model-free RL policy of the disembodied manipulator is learned from interacting with articulated objects. The QP-based controller then optimizes the high-dimensional joint-space actions to approximate the learned skill dynamics by synchronizing trajectories of the robot's end-effector and the disembodied manipulator. We evaluate the category-level generalizability of the learned skill policy over objects and achieve an average success rate of 74\% on the training set and 51\% on the test set. For future work, we would like to extend the generalizability of our method to more challenging conditions, including more complex tasks and robots with constraints or inaccuracies in dynamics modeling.




\bibliographystyle{unsrt}
\bibliography{root}

\end{document}